\documentclass[aos]{imsart}

\RequirePackage{amsthm,amsmath,amsfonts,amssymb}
\RequirePackage[numbers,sort&compress]{natbib}
\RequirePackage[colorlinks,citecolor=blue,urlcolor=blue]{hyperref}
\RequirePackage{graphicx}

\startlocaldefs
\theoremstyle{plain}

\newtheorem{theorem}{Theorem}[section]
\newtheorem{lemma}[theorem]{Lemma}

\theoremstyle{remark}
\newtheorem{definition}[theorem]{Definition}
\newtheorem{assumption}{Assumption}
\newtheorem{notation}{Notation}

\newtheorem*{remark}{Remark}

\endlocaldefs

\begin{document}

\begin{frontmatter}
\title{A Robbins--Monro Sequence That Can Exploit Prior Information For Faster Convergence}

\begin{aug}
\author{\fnms{Siwei}~\snm{Liu}\ead[label=e1]{sl2049@cam.ac.uk}\orcid{0009-0002-4176-3381}},
\author{\fnms{Ke}~\snm{Ma}\ead[label=e2]{km834@cam.ac.uk}\orcid{0000-0002-8499-4529}}
\and
\author{\fnms{Stephan M.}~\snm{Goetz}\ead[label=e3]{smg84@cam.ac.uk}\orcid{0000-0002-1944-0714}}
\address{
University of Cambridge\\ \printead{e1,e2,e3}}
\end{aug}

\begin{abstract}
We propose a new way to improve the convergence speed of the Robbins--Monro algorithm by introducing prior information about the target point into the Robbins--Monro iteration. We achieve the incorporation of prior information without the need of a---potentially wrong---regression model, which would also entail additional constraints.
We show that this prior-information Robbins--Monro sequence is convergent for a wide range of prior distributions, even wrong ones, such as Gaussian, weighted sum of Gaussians, e.g., in a kernel density estimate, as well as bounded arbitrary distribution functions greater than zero.
We furthermore analyse the sequence numerically to understand its performance and the influence of parameters. The results demonstrate that the prior-information Robbins--Monro sequence converges faster than the standard one, especially during the first steps, which are particularly important for applications where the number of function measurements is limited, and when the noise of observing the underlying function is large. We finally propose a rule to select the parameters of the sequence. 
\end{abstract}



\end{frontmatter}

\section{Introduction}
The Robbins--Monro (RM) iteration was first proposed by Herbert Robbins and Sutton Monro as a general stochastic sequence, which can be used for almost-surely (a.s.{}) converging root finding of noisy function observations \cite{robbins1951stochastic}. Due to its robustness to variability and fast convergence, it has become one of the most widely used algorithms in applied mathematics and related disciplines, such as in dealing with large data sets as well as machine learning \cite{bottou2018optimization, moulines2011non, zhang2004solving}, weather forecasting \cite{iooss2021robust}, engineering \cite{marti2003stochastic, lai2003stochastic}, medicine, as well as neuroscience \cite{goetz2019statistical,wang2022analysis,li2022detection}. This sequence, also often marketed under the label \textit{stochastic approximation}, can estimate the point $x_\textrm{t}$ where a function $f(x)$ reaches a certain function value $f(x_\textrm{t}) = y_\textrm{t}$, e.g., a root. Function $f(x)$ is unknown but 
can be observed through and unbiasedly estimated by $y(x)$, which further contains a stochastic component \cite{robbins1951stochastic}. Thus, all observations are inherently variable or subject to noisy measurements. The sequence can further be modified to find the root of an approximation of the gradient of a function, for example, through a converging finite-difference representation, and result in a stochastic optimisation method for finding the minima of variably or noisy functions.

In the following, let $y(x)=f(x) + e_k$, expectation $E[y(x)]=f(x)$, and $e_k$ be a bounded variability or error term with 0 mean, changing with every function evaluation $k$. Denote the starting point of this algorithm by $x_1$ and the target point by $x_\textrm{t}$, where $f(x_\textrm{t}) = y_\textrm{t}$. The iteration of the RM sequence follows
$$
x_{i+1}=x_i-s_i(y_i-y_\textrm{t}).
$$

The abbreviation of $y_i$ denotes the function values $y(x_i)$, and the sequence $s_i$ represents the (decreasing) step size or gain. The latter translates a deviation of the observation or measurement of $y_i$ from the target value $y_i$ into a new $x$, i.e., input value to be tested. As was previously found, the selection of a good step size sequence is closely related to the local steepness of the function $f$. For the RM algorithm to be a.s.\ convergent, the step size sequence must further satisfy $\sum_{i=1}^\infty s_i=\infty$, $\sum_{i=1}^\infty s_i^2 < \infty$. Following the harmonic and Basel series theory of Mengoli \cite{Mengoli1650} and Euler \cite{Euler1744}, that means that $s_i$ must---in the long run---be decreasing approximately on an order faster than $i^{-0.5}$ and slower than or equal to $i^{-1}$. Here, $\sum_{i=1}^\infty s_i=\infty$ guarantees that the RM iteration will be able to converge to $x_\textrm{t}$ no matter how far $x_1$, $x_\textrm{t}$ are from each other, and $\sum_{i=1}^\infty s_i^2 < \infty$ guarantees that $\lim_{i \to \infty}s_i=0$, and hence the algorithm can be convergent. Robbins and Monro proved that this sequence converges to $x_\textrm{t}$ in quadratic mean ($E[(x_i-x_\textrm{t})^2]=0$) when $y(x)$ is bounded \cite{robbins1951stochastic}.

After the RM algorithm was proposed, much work has been done to strengthen the convergence result of the RM algorithm. Blum \cite{blum1954approximation} as well as Robbins and Siegmund \cite{robbins1971convergence} demonstrated that the RM algorithm is convergent when $f(x)$ may by itself be unbounded in the strict sense but is bounded by two linear functions. Kallianpur derived that the RM algorithm converges with probability 1 \cite{kallianpur1954note}. Kushner and Yin \cite{kushnerstochastic}, Borkar \cite{borkar2009stochastic}, as well as Ljung et al.\ \cite{ljung2012stochastic} strengthened this convergence result from the perspective of dynamical system theory. In their analysis, the RM algorithm is viewed as a discrete-time stochastic process. Each iteration of the algorithm is treated as a step in the evolution of a dynamical system. Having found an appropriate Lyapunov function, they demonstrated that the trajectories of the dynamical system converge to a stable equilibrium point ($x_\textrm{t}$) by showing that the Lyapunov function decreases over time. Moreover, the asymptotic properties of the RM sequence were also investigated. Chung \cite{chung1954stochastic}, Hodges \cite{hodges1956two}, and Sacks \cite{sacks1958asymptotic} proved that $\sqrt{s_i}(x_i-x_\textrm{t})$ is asymptotically normal. Moreover, Chung considered the asymptotic properties of the higher-order moments of ${x_i-x_\textrm{t}}$ \cite{chung1954stochastic}. Fabian \cite{fabian1968asymptotic} later simplified the proof of Chung \cite{chung1954stochastic}.

Research has studied many modifications and generalisations of the RM sequence. A continuous variant of the RM sequence reads 
$$
\begin{aligned}
&\frac{\textrm{d}}{\textrm{d}t} X(t) = -s(t)Y(t,X(t)),\\
&\int_0^{\infty} \!\! s(t)\, \textrm{d}t = \infty, \\
&\int_0^{\infty}\!\! s^2 (t) \,\textrm{d}t < \infty.
\end{aligned}
$$

Driml and Nedoma demonstrated that the RM algorithm is convergent when $Y(t,x)$ is monotonic in $x$ and $Y(t,x) = M(x) + h(t)$ where $h(t)$ is an ergodic process with zero mean \cite{driml1960stochastic}. However, when extending the one-dimensional continuous case to the multidimensional continuous case, Hans and Spacek showed that many results from one dimension do not hold for multidimensional functions \cite{hans1960random}. 

Blum first investigated the convergence of a multidimensional discrete RM algorithm \cite{blum1954multidimensional}. In his assumptions, the sequence would move in a direction in which $\|x_n -x_\textrm{t}\|$ decreased. Wei \cite{wei1987multivariate} and Krishnaiah \cite{krishnaiah1969simultaneous} investigated the asymptotic properties of the multidimensional discrete RM case. Ruppert modified Blum’s multidimensional RM iteration and proposed a new sequence with a different stepping rule that aims at decreasing $\|f\|^2$ step by step \cite{ruppert1985newton}.

Moreover, Joseph \cite{joseph2004efficient} and Wu \cite{wu1985efficient} examined the RM iteration for binary data. Xiong and Xu further generalised the results of the binary RM into multiple dimensions \cite{xiong2018efficient}.

Early in the study of the properties of the RM algorithm, its often slow convergence speed and instability became an important topic. For higher stability of the RM algorithm while keeping its convergence speed, Toulis et al.\ proposed a proximal version for the RM algorithm to allow better stability by penalising sudden large steps the further the iteration progresses  \cite{toulis2021proximal}. Moreover, intensive research aimed to increase the RM algorithm's convergence speed. Dvoretzky found a sequence of step sizes ${s_i}$ that minimises $E[(x_i -x_\textrm{t})^2]$ after some fixed number of iterations \cite{dvoretsky1955stochastic}. Kesten first proposed a widely-used intuitive and accordingly repeatedly re-suggested method for accelerating the convergence of the RM algorithm \cite{kesten1958accelerated}. His improvement reduces the step size sequence $s_i$ only, if $x_i - x_{i-1}$ and thus the deviation of the function observation from the target value $y_i - y_\textrm{t}$ has the opposite sign compared to the previous one $x_{i-1} - x_{i-2}$ or $y_{i-1} - y_\textrm{t}$; otherwise, the step size $s_i$ will remain the same as $s_{i-1}$. Venter modified the iteration of the standard RM algorithm and included information about the local derivative based on an estimate of $f^{\prime}(x_\textrm{t})$ to improve the rate of convergence and the asymptotic variance of the Robbins--Monro procedure \cite{venter1967extension}. Still, the convergence speed of the RM iteration shows great potential. 

The convergence speed in the first steps of the iteration can determine the overall speed and is furthermore important in many practical applications where only a few function evaluations are possible \cite{gotz2011ptms59}. Farrell worked on achieving a statistical grasp on the residual deviation of the iteration in the form of $P(|x_{i+1}-x_\textrm{t}| \leq D) \geq 1 - 2\gamma$ after a finite number of steps \cite{farrell1962bounded, farrell1959sequentially}. Chung derived lower and upper bounds for $E[|x_n-x_\textrm{t}|^r]$ and $E[(x_n-x_\textrm{t})^r]$ for all $n$ \cite{chung1954stochastic}. Kallianpur further demonstrated that $E[(x_n-x_\textrm{t})^2] = \mathcal{O}(\log\,{}n)^{-v}$ for some constant $v$ in his convergence proof of RM algorithm \cite{kallianpur1954note}. Polyak used the Lyapunov function to illustrate the rate of convergence of RM \cite{polyak1976convergence}. Based on these results, Sielken \cite{sielken1973some} as well as Stroup and Braun \cite{stroup1982new} investigated stopping rules for scalar nonlinear $f(x)$. Such rules also promise to shine light on the early convergence speed of RM. Yin investigated stopping rules for the multidimensional case \cite{yin1988stopped}. However, these rules are all in the form of limits, which are not explicit enough for the first few steps. Wada et al.\  gave an intuitive equation describing the stopping time of the RM algorithm for different target accuracy and initial point $x_1$ for linear stochastic approximation \cite{wada2010stopping}. Their equation can estimate the RM algorithm’s accuracy also for only limited number of iterations. Wada and Fujisaki generalised the results to nonlinear stochastic approximation \cite{wada2015stopping}. However, by now, there is no further method that can significantly speed up the convergence of the RM algorithm in the first few steps. It is often simply assumed and accepted that the sequence $x_i$ is typically far from the target $x_\textrm{t}$ when $i$ is small \cite{glynn1992asymptotic}. 

Stochastic approximation methods (and particularly Robbins--Monro) can further solve stochastic optimisation problems (which are equivalent to finding the root of the derivative $f' =0$ for an increasing derivative $f'$), which is a major application of the Robbins--Monro sequence. Previous research could demonstrate that if $f(x)$ is twice continuously differentiable and strongly convex, the algorithm can attain the asymptotically optimal convergence rate $\mathcal{O}(1/n)$, that is, $E[f(x_n)-f(x_\textrm{t})] = \mathcal{O}(1/n)$ \cite{spall2005introduction, nemirovski2009robust}. In the general convex case of $f(x)$, the asymptotically optimal convergence rate can only achieve $\mathcal{O}(1/\sqrt{n})$. In the long run, the order of this convergence rate appears already the best possible \cite{nemirovski1978cezari, nemirovskij1983problem}. 

In this paper, we intend to accelerate the convergence of the RM iteration, particularly early on, in the first steps. In many scientific and practical problems, information about the convergence point is known on a statistical level, typically the distribution is known from previous usage on similar cases or can be roughly estimated before starting the approximation process. Historical data can, for example, provide such prior information for numerical weather and climate problems. Known population data---likely estimated with the very same RM procedure---may in turn serve for threshold detection in medicine, neuroscience, or psychophysics \cite{wassermann2002variation, ma2023correlating, treutwein1995adaptive, psychophysical}. As for the problem of slow convergence speed mentioned above, the inclusion of such prior knowledge should improve the convergence speed. This article will introduce a sequence that expands Robbins--Monro to entire distributions and can exploit prior information in the iteration to speed up the convergence. 

This text is structured as follows: Section \ref{s2} introduces the new sequence that incorporates the prior information about $x_\textrm{t}$. Sections \ref{s3} -- \ref{s4} provide two methods to prove the convergence of our new sequence with linear $f(x)$ and Gaussian prior. The convergence proof of nonlinear $f(x)$ with Gaussian and practically arbitrary prior follows in Sections \ref{s5} and \ref{s6}. The general convergence allows for an alternative proof as coarsely sketched in Section \ref{s7}. We numerically analysed the sequence for demonstrating its performance and better understanding of its parameters and summarised the results in Section \ref{s8}. Section \ref{s9} concludes this paper.

\section{Integration of prior information into the Robbins--Monro sequence} \label{s2}
\subsection{Formalism}
\begin{notation}
$\mathcal{N}(x \,|\, \alpha, \beta^2)$ is the normal distribution with $\alpha$ mean and $\beta^2$ variance.\\ 
\end{notation}
For an integration of prior information about the convergence point, we turn the RM iteration into a sequence of statistical distributions. We intuitively interpret the conventional RM iteration as the solution of a maximum likelihood estimator without any prior information evaluated in each step. For the case of near-Gaussian likelihood, the maximum and the spread would fully determine the shape. With convergence, the distribution would pointwise converge so that the spread in the form of the standard deviation decreases on the order of $1/i$. Thus, the best estimate of $x_{i+1}$ with the information from the previous measurement would obviously be
$$
x_{i+1}=\operatorname{argmax}_x P(x_{i+1} \mid x_i, s_i, \mathrm {Model}),
$$
where
$$
P(x_{i+1} \mid x_i, s_i, \mathrm {Model})=  \mathcal{N}\Big(x \,\Big|\, x_i-s_i(y_i-y_\textrm{t}), c_i^2\Big).
$$
Note that $c_i = c_0/i$, $c_0$ is a positive constant. Moreover, $\mathcal{N}\big(x \,|\, x_i-s_i(y_i-y_\textrm{t}), c_i^2 \big)$ is a normal distribution with $x_i-s_i(y_i-y_\textrm{t})$ mean and $c_i^2$ variance.

If we have some prior information of $x_\textrm{t}$, we can use Bayes’ Theorem to derive
$$
P(x_\textrm{t} \mid x_{i+1}) \propto P_{x_\textrm{t}}(x) P(x_{i+1} \mid x_i, s_i, \mathrm {Model}),
$$
where $P_{x_\textrm{t}}(x)$ is the prior information describing the probability distribution of $x_\textrm{t}$. We want to proceed in the RM sequence containing the prior information with the best estimate of $x_\textrm{t}$ as the next step. That is, we want to choose that $x_{i+1}$ which maximizes $P(x_\textrm{t} \mid x_{i+1})$. Therefore,
\begin{align}
x_{i+1}=\operatorname{argmax}_x \!\left(P_{x_\textrm{t}}(x) \cdot \mathcal{N}\Bigl(x\, \Big|\, x_i-s_i(y_i-y_\textrm{t}) , c_i^2\Bigr)\right).\label{def}
\end{align}

For brevity, we will in the following name $P_{x_\textrm{t}}(x)$ the prior distribution and $\mathcal{N}\big(x \mid x_i-s_i(y_i-y_\textrm{t}), c_i^2 \big)$  the RM distribution. Further, we will denote the entire sequence prior-information Robbins--Monro sequence here. Note that when the prior distribution in Eq.\ (\ref{def}) is a uniform prior (noninformative prior), the prior-information Robbins--Monro sequence will coincide with the standard RM sequence. When the prior information is known and relatively accurate, it should significantly accelerate the convergence speed, especially for the first steps when $x_i$ is far from $x_\textrm{t}$. 

\subsection{Interpretation}
The prior distribution can be practically any positive bounded continuous probability distribution. In the following analysis, we will first assume the prior distribution to be a normal distribution and then generalise the results derived from the normal prior distribution to any practical prior distribution (non-zero to allow for any outcome, bounded, and finite weight under the curve (obviously typically 1) so that the RM distribution can eventually outgrow its maximum). Ideally, the prior distribution should (relatively) accurately describe the distribution of $x_\textrm{t}$ expected in the specific problem. However, the prior distribution could also be wrong. If it does not suppress the right outcome, it should still allow---albeit typically slower---convergence. The results in our following analysis demonstrate that our algorithm will still be convergent even if the prior information is wrong if it is non-zero and finite in the relevant ranges, specifically a large enough neighbourhood of the actual root of the function.

Note that the mean of the RM distribution in step $i$ is the value $x_{i+1}$ of the standard RM algorithm, and the variance of the RM distribution decreases with the number of iterations. The product of the prior distribution $P_{x_\textrm{t}}(x)$ with the RM distribution $\mathcal{N}\bigl(x\, \big|\, x_i-s_i(y_i-y_\textrm{t}) , c_i^2\bigr)$ allows the former to pull the maximum closer to the maximum of the prior, i.e., where prior information would expect the root of the function $f(x)$. However, whereas the prior distribution stays the same throughout the iteration, the RM distribution  becomes narrower and sharper with every step.  Furthermore, its maximum grows beyond any value of the bounded prior as well as also any product of the maximum of the prior and a value of the RM distribution outside a neighbourhood of the RM distribution maximum. This neighbourhood likewise decreases on a $1/i$ trend. Thus, the maximum of the RM distribution eventually dominates the argmax operator. From the perspective of the sharper-growing RM distribution, the prior distribution practically degrades into a constant, i.e., a naive prior, in the shrinking neighbourhood of the RM distribution maximum after a sufficient number of iterations.
Thus, the prior distribution exerts its influence, particularly during the early steps, until the RM distribution turns into a dominant sharp peak.

\section{Convergence for linear function $f$ and Gaussian prior} \label{s3}

\begin{notation}\label{n2} In all the following proof, $\prod_{u=m}^{n} Q(u) =1$ if $m>n$ for any arbitrary function $Q$.
\end{notation}

\begin{definition}\label{def1}
Let $x_1^{\mathrm{s}}, x_2^{\mathrm{s}},...$ be the standard Robbins--Monro sequence. $s_1, s_2,...$ are step size gains for the standard Robbins--Monro sequence. Then, the standard Robbins--Monro sequence follows 
$$
x_{i+1}^{\mathrm{s}}=x_i^{\mathrm{s}}-s_i(y_i^{\mathrm{s}}-y_\textrm{t}).
$$

In the Robbins--Monro sequence, the target point we want to find is $x_\textrm{t}$. Assume $f(x_\textrm{t})=y_\textrm{t}$, and $\sum_{i=1}^\infty s_i=\infty$, $\sum_{i=1}^\infty s_i^2 < \infty$. Moreover, $y_i^{\mathrm{s}}=y(x_i^{\mathrm{s}})$ is the measurement of $f(x)$ at $x_i^{\mathrm{s}}$ with $y_i^{\mathrm{s}}= f(x_i^{\mathrm{s}})+\varepsilon_{i}^{\mathrm{s}}$, $\varepsilon_{i}^{\mathrm{s}},...$ are bounded independent random variables with 0 mean. \\
\end{definition}

\begin{definition}
Let $x_1, x_2,...$ be the prior-information Robbins--Monro sequence, and $s_1, s_2,...$ of Definition \ref{def1} be the step sizes also for the prior-information Robbins--Monro sequence. Then, the prior-information Robbins--Monro sequence follows 
$$
x_{i+1}=\operatorname{argmax}_x\!\left(P_{x_\textrm{t}}(x) \cdot \mathcal{N}\Big(x \Big| x_i-s_i(y_i-y_\textrm{t}), c_i^2\Big)\right),
$$
where $c_i =c_0/i$, $c_0$ is a positive constant. $P_{x_\textrm{t}}(x)$ describes the assumed probability distribution of $x_\textrm{t}$. Denote $P_{x_\textrm{t}}(x)$ as prior distribution, and designate $\mathcal{N}\big(x \mid x_i-s_i(y_i-y_\textrm{t}), c_i^2\big)$ the RM distribution. $y_i=y(x_i)$ is the measurement of $f(x)$ at $x_i$. Moreover, $y_i= f(x_i)+\varepsilon_{i}$ with $\varepsilon_{i},...$ are bounded independent and identically distributed (i.i.d.) random variables with 0 mean and $d^2$ variance ($d<\infty$). The target point we want to find is likewise $x_\textrm{t}$, $f(x_\textrm{t})=y_\textrm{t}$.
\\
\end{definition}
\begin{assumption} If $f(x)$ is linear, w.l.o.g., assume $f(x)=ax$, $a>0$.
\end{assumption}
\begin{assumption}
Assume $x_1 = x_1^{\mathrm{s}}$ in the following proof.
\end{assumption}
\begin{assumption} \label{a3}
If $f(x)$ is a nonlinear function, then assume $f(x)$ is continuously differentiable, positive, and bounded per  $M \geq f^{\prime} \geq m > 0$.
\end{assumption}
\begin{assumption}
Assume w.l.o.g.\ that the variance of the RM distribution follows $1/i^2$ in the following proof.
\end{assumption}
\begin{assumption}
$\{\varepsilon_i\}$ and $\{\varepsilon_i^{\mathrm{s}}\}$ are symmetric about 0. With this assumption, $\varepsilon_i$ will be equal to  $-\varepsilon_i$, $\forall i \in \mathbb{Z}^{+}$.
\end{assumption}
\textbf{By default, Assumptions 1--5 hold for the proof below.}

Note that we do not require the prior distribution to correctly describe the possible values of $x_\textrm{t}$.
\begin{figure}
    \centering
    \includegraphics[scale=0.88]{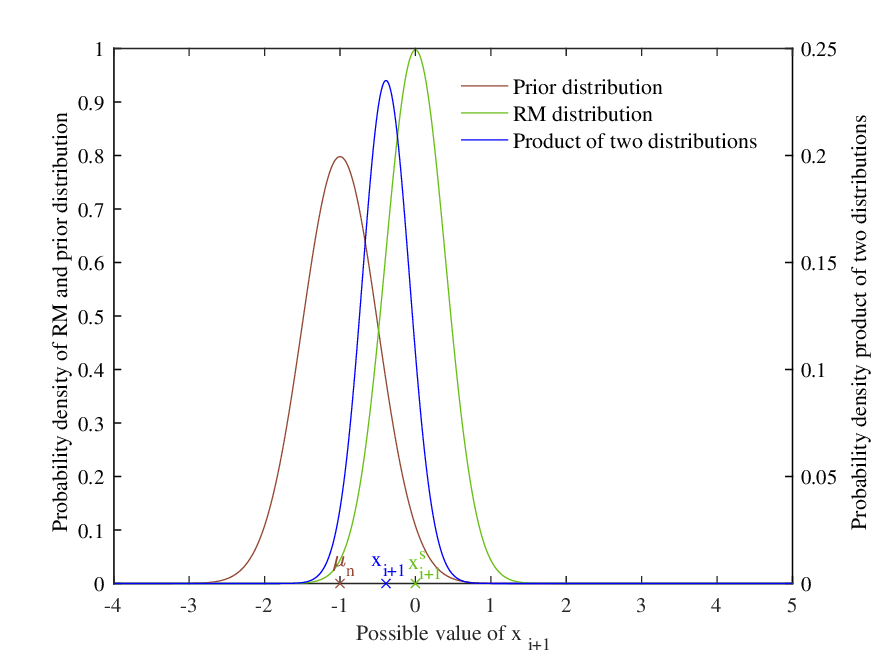}
    \caption{Illustration of the prior-information Robbins--Monro sequence with the two contributions of prior distribution and Robbins--Monro distribution combined to provide the a-posteriori distribution.}
    \label{fig0.5}
\end{figure}

\begin{lemma}\label{lem:important}
When $f(x)=ax$ is linear ($a>0$), for any finite $h \in \mathbb{Z}^{+},$
$$
\begin{aligned}
\prod_{k=h}^\infty (1-a s_k) = 0 \quad a.s.
\end{aligned}
$$
\begin{proof}
\begin{align}
x_{i+1}^{\mathrm{s}} &=x_i^{\mathrm{s}} -s_i (a(x_i^{\mathrm{s}}-x_\textrm{t})+\varepsilon_{i}^{\mathrm{s}})\notag\\
x_{i+1}^{\mathrm{s}}-x_\textrm{t} &=(1-as_i)(x_i^{\mathrm{s}}-x_\textrm{t}) +s_i \varepsilon_{i}^{\mathrm{s}}\notag\\
&=\prod_{k=1}^{i}(1-as_k)(x_1-x_\textrm{t})+\sum_{k=1}^{i}s_k\varepsilon_{k}^{\mathrm{s}}\prod_{u=k+1}^{i} (1-as_u)\label{eq1.5}
\end{align}

Here, we write 
$$\sum_{k=1}^{i-1}s_k\varepsilon_{k}^{\mathrm{s}}\prod_{u=k+1}^{i} (1-as_u)+ s_i\varepsilon_{i}^{\mathrm{s}}$$ 
into 
$$\sum_{k=1}^{i}s_k\varepsilon_{k}^{\mathrm{s}}\prod_{u=k+1}^{i} (1-as_u)$$
with Notation \ref{n2} for simplification. We will frequently use this abbreviation in the following proofs.

Note that $x_1^{\mathrm{s}}=x_i$. Since the standard Robbins--Monro algorithm is convergent a.s.\ and $E(\varepsilon_{k}^{\mathrm{s}})=0$ $\forall k$, then
$$
\begin{aligned}
&\lim_{i \to \infty}E \big[x_{i+1}^{\mathrm{s}}-x_\textrm{t} \big] \\
=&\lim_{i \to \infty} E \Big[\prod_{k=h}^{i}(1-as_k)(x_h^{\mathrm{s}}-x_\textrm{t})+\sum_{k=h}^{i}s_k\varepsilon_{k}^{\mathrm{s}}\prod_{u=k+1}^{i} (1-as_u) \Big]\\
=&\prod_{k=h}^{\infty}(1-as_k)(x_h^{\mathrm{s}}-x_\textrm{t}) =0
\end{aligned}
$$
for all finite $h \in \mathbb{Z}^{+}$. 

Moreover, $x_h^{\mathrm{s}}-x_\textrm{t} \neq 0$ a.s.\ since the value of $x_h^{\mathrm{s}}$ is influenced by $h-1$ continuous random variables $\varepsilon_{1}^{\mathrm{s}},\varepsilon_{2}^{\mathrm{s}},...,\varepsilon_{h-1}^{\mathrm{s}}$. Therefore,
\begin{align*}
 \prod_{k=h}^{\infty}(1-as_k) = 0 \quad a.s.
\end{align*}
for all finite $h \in \mathbb{Z}^{+}$.
\end{proof}
\end{lemma}

\begin{lemma}\label{lem2}
When  $f(x)=ax$ is linear ($a>0$), $$\sum_{k=1}^{\infty} s_k^2 d^2 \prod_{u=k+1}^{\infty}(1-a s_u)^2  = 0 \quad a.s.$$ 
Note that $d^2$ is the variance of $\{\varepsilon_i\}$.
\begin{proof}
Here, we want to further consider a standard RM sequence that satisfies $\varepsilon_k^{\mathrm{s}},$... are i.i.d with 0 mean and $d^2$ variance. Note that this assumption will not change the convergence of the standard RM sequence. Calculate the square of Eq.\ (\ref{eq1.5}) and take the expectation to get
\begin{align}
E[(x_{i+1}^{\mathrm{s}}-x_\textrm{t})^2]&=\prod_{k=1}^i(1-a s_k)^2(x_1-x_\textrm{t})^2 \label{eq2}\\
&+\sum_{k=1}^i s_k^2 d^2 \prod_{{u=k+1 }}^i(1-a s_u)^2 \label{eq3}
\end{align}
since $\varepsilon_{i}^{\mathrm{s}},$... are i.i.d.\ random variables with 0 mean and $d^2$ variance. As the standard RM sequence is a.s.\ convergent, 
\begin{align*}
\lim_{i \to \infty}E[(x_{i+1}^{\mathrm{s}}-x_\textrm{t})^2]  = 0 \quad a.s.
\end{align*}
\\
Since Terms $(\ref{eq2}), (\ref{eq3}) \geq 0$, then Terms $(\ref{eq2}), (\ref{eq3}) \stackrel{i\to \infty}{\longrightarrow} 0$ a.s.{},
$$
\sum_{k=1}^{\infty} s_k^2 d^2 \prod_{u=k+1 }^{\infty}
(1-a s_u)^2 =0 \quad a.s.
$$
\end{proof}
\end{lemma}

\begin{theorem}\label{thm1}
When the prior distribution is normal and $f(x)=ax$ is linear ($a>0$), the prior-information Robbins--Monro sequence is generated by $$x_{i+1}=\operatorname{argmax}_x\left(\mathcal{N} ( x \mid \mu_n, \sigma_n^2) \cdot \mathcal{N}\Big(x \Bigm| x_i-s_i(a(x_i-x_\textrm{t})+\varepsilon_i), \frac{1}{i^2}\Big)\right),$$ where $ \mu_n, \sigma_n$ are  constants, $ \sigma_n >0$. Then, 
$$\lim _{i \rightarrow \infty}x_{i+1}= x_\textrm{t} \quad a.s.$$
Therefore, the prior-information Robbins--Monro sequence is convergent a.s.\ for linear $f(x)$ and normal prior distribution.
\begin{proof}
\begin{align*}
x_{i+1} &=\operatorname{argmax}_x\!\left(\mathcal{N}(x \mid \mu_n, \sigma_n^2) \cdot \mathcal{N}\Big(x \Bigm| x_i-s_i(a(x_i-x_\textrm{t})+\varepsilon_i), \frac{1}{i^2}\Big)\right) \\
& =\operatorname{argmax}_x\!\left(-\frac{(x-\mu_n)^2}{2 \sigma_n^2}-\frac{i^2(x-x_i+s_i(a(x_i-x_\textrm{t})+\varepsilon_i))^2}{2}\right)\\
&=\operatorname{argmax}_x g(x) 
\end{align*}
\\
Since
\begin{align*}
g^{\prime \prime}(x) =-\frac{1}{\sigma_{n}^2 }-i^2<0,\mathrm{ \ we \ know \ that \ } g^{\prime}(x_{i+1})=0. \\
\end{align*}
\\
We can have
\begin{align*}
x_{i+1} &=\frac{ \mu_n+i^2 \sigma_n^2 (1-a s_i ) x_i+i^2 \sigma_n^2 s_i (a x_\textrm{t}+\varepsilon_i )}{i^2 \sigma_n^2+1} \\
x_{i+1}-x_\textrm{t} &=\frac{i^2 \sigma_n^2 (1-a s_i )}{i^2 \sigma_n^2+1} (x_i-x_\textrm{t} )+\frac{ \mu_n-x_\textrm{t} }{i^2 \sigma_n^2+1}+\frac{i^2 \sigma_n^2 s_i \varepsilon_i}{i^2 \sigma_n^2+1} 
 \end{align*}
 \\
Hence, we can iteratively form
\begin{align}
\lim_{i \to \infty} x_{i+1}-x_\textrm{t} & =\prod_{k=h}^{\infty} \frac{k^2 \sigma_n^2 (1-a s_k )}{k^2 \sigma_n^2+1} (x_h-x_\textrm{t} )\label{eq4}\\
&+\sum_{k=h}^{\infty} \frac{\mu_n-x_\textrm{t} }{k^2 \sigma_n^2+1} \cdot \prod_{u=k+1}^{\infty} \frac{u^2 \sigma_n^2 (1-a s_u )}{u^2 \sigma_n^2+1} \label{eq4.1}\\
& +\sum_{k=h}^{\infty} \frac{k^2 \sigma_n^2 s_k \varepsilon_k}{k^2 \sigma_n^2+1} \cdot \prod_{u=k+1}^{\infty} \frac{u^2 \sigma_n^2 (1-a s_u )}{u^2 \sigma_n^2+1}, \label{eq4.2}
\end{align}
where $h$ is a given large number that satisfies $s_k< 1/a,$ $\forall k \geq h$. Such $h$ exists since $\sum s_i^2<\infty$, $ s_i \stackrel{i\to \infty}{\longrightarrow} 0$.\\
\\
By Lemma \ref{lem:important},
$$\prod_{k=h}^{\infty}(1-as_k) =0 \quad a.s.$$
\\
Therefore, 
$$
0 \leq \left|\prod_{k=h}^{\infty} \frac{k^2 \sigma_n^2 (1-a s_k )}{k^2 \sigma_n^2+1} (x_h-x_\textrm{t} ) \right|
\leq  \left|\prod_{k=h}^{\infty}(1-as_k)(x_h-x_\textrm{t}) \right|=0 \quad a.s.{},
$$
and Term (\ref{eq4}) = 0 a.s.\\
\\
For Term (\ref{eq4.1}), since $ \sum_{k=1}^{\infty} 1/k^2 < \infty$, for any infinitesimal $p >0$, there exists $Z \in \mathbb{N}$ such that 
$$\left| \sum_{k=Z}^{\infty} \frac{\mu_n-x_\textrm{t}}{k^2 \sigma_n^2}\right| < p,$$
w.l.o.g., assume $Z>h$. Therefore, \\
$$
0 \leq \left|\sum_{k=Z}^{\infty} \frac{\mu_n-x_\textrm{t} }{k^2 \sigma_n^2+1} \cdot \prod_{u=k+1}^{\infty} \frac{u^2 \sigma_n^2 (1-a s_u )}{u^2 \sigma_n^2+1}\right| \leq \left|\sum_{k=Z}^{\infty} \frac{\mu_n-x_\textrm{t} }{k^2 \sigma_n^2+1}\right| \leq \left|\sum_{k=Z}^{\infty} \frac{\mu_n-x_\textrm{t} }{k^2 \sigma_n^2} \right| < p.
$$
\\
Hence,
\begin{align}
\sum_{k=Z}^{\infty} \frac{\mu_n-x_\textrm{t} }{k^2 \sigma_n^2+1} \cdot \prod_{u=k+1}^{\infty} \frac{u^2 \sigma_n^2 (1-a s_u )}{u^2 \sigma_n^2+1}=0.\label{sa}
\end{align}
Moreover, when $h \leq k \leq Z$, following the same idea in proving in Term (\ref{eq4}) $=0$, we can get
$$
\prod_{u=k+1}^{\infty} \frac{u^2 \sigma_n^2 (1-a s_u )}{u^2 \sigma_n^2+1}=0 \quad a.s.{},
$$
and 
\begin{align}
\sum_{k=h}^{Z-1} \frac{\mu_n-x_\textrm{t} }{k^2 \sigma_n^2+1} \cdot \prod_{u=k+1}^{\infty} \frac{u^2 \sigma_n^2 (1-a s_u )}{u^2 \sigma_n^2+1}=0 \quad a.s.\label{sb}
\end{align}
Therefore, adding up Eq. (\ref{sa}) and Eq. (\ref{sb}), we can get Term (\ref{eq4.1}) = 0 a.s.\\
\\
Calculate the square of Term (\ref{eq4.2}) and take the expectation gets us to
\begin{align}
E [ (\ref{eq4.2})^2 ]  =\sum_{k=h}^{\infty} \left(\frac{k^2 \sigma_n^2 s_k}{k^2 \sigma_n^2+1} \right)^2 d^2 \cdot \prod_{u=k+1}^{\infty} \left(\frac{u^2 \sigma_n^2 (1-a s_u )}{u^2 \sigma_n^2+1} \right)^2 \label{eq6}
\end{align}
since $\varepsilon_k, \varepsilon_{k+1},...$ are i.i.d random variables with 0 mean and $d^2$ variance.\\
\\
By Lemma \ref{lem:important}, when $k=1,2,3,...,h-1$,
$$
\prod_{u=k+1}^{\infty} (1-as_u) =0 \quad a.s.
$$
Hence,
$$
\sum_{k=1}^{h-1} s_k^2 d^2\prod_{u=k+1}^{\infty} (1-as_u)^2 =0 \quad a.s.
$$
\\
Moreover, with Lemma \ref{lem2}, we can get 
$$0 \leq|(\ref{eq6})| \leq \left|\sum_{k=h}^{\infty} s_k^2 d^2\prod_{u=k+1}^{\infty} (1-as_u)^2\right|  = 0 \quad a.s.
$$
Therefore Eq.\ (\ref{eq6})  = Term\ (\ref{eq4.2}) $= 0$ a.s.\\
\\
Accordingly $$\lim_{i \to \infty}x_{i+1} \stackrel{a.s.}{=} x_\textrm{t} ,$$
and the prior-information Robbins--Monro sequence is convergent a.s.\ for linear $f(x)$ and normal prior distribution.
\end{proof}
\end{theorem}

\begin{remark}
Note that if the prior distribution $\mathcal{N} ( x \mid \mu_n, \sigma_n^2)$ does not correctly describe the possible distribution of $x_\textrm{t}$, the prior-information Robbins--Monro sequence is still convergent a.s. This remark holds for all the following convergence proofs.
\end{remark}

\section{An alternative convergence proof for a linear function $f$ and Gaussian prior} \label{s4}
We will also provide an alternative idea to prove the a.s.\ convergence of the prior-information Robbins--Monro sequence.
\begin{assumption} \label{a5}
w.l.o.g, assume $x_\textrm{t}=y_\textrm{t}=0$. \\
\end{assumption}
\begin{theorem}\label{thm1.11}
When the prior distribution is normal and $f(x)=ax$ is linear ($a>0$), the prior-information Robbins--Monro sequence is convergent a.s.\ when Assumption \ref{a5} holds.
\begin{proof}
In this proof, we will construct a standard RM sequence $\{x_i^{\mathrm{s}}\}$ that has the same limit as our prior-information Robbins--Monro sequence $\{x_i\}$ to show that both $\{x_i\}$ and $\{x_i^{\mathrm{s}}\}$ converges to $x_\textrm{t}$. \\
\\
Follow the same idea in Lemma \ref{lem:important} and Theorem \ref{thm1}, we can get\\
$$
\begin{aligned}
& \lim_{i \to \infty} x_{i+1}=\prod_{k=h}^{\infty} \frac{k^2 \sigma_n^2\left(1-a s_k\right)}{k^2 \sigma_n^2+1} x_h+\sum_{k=h}^{\infty} \frac{\mu_n+k^2 \sigma_n^2 s_k \varepsilon_k}{k^2 \sigma_n^2+1} \cdot \prod_{u=k+1}^{\infty} \frac{u^2 \sigma_n^2\left(1-a s_u\right)}{u^2 \sigma_n^2+1} \\
& \lim_{i \to \infty} x_{i+1}^{\mathrm{s}}=\prod_{k=h}^{\infty}\left(1-a s_k\right) x_h^{\mathrm{s}}+\sum_{k=h}^{\infty} s_k \varepsilon_k^{\mathrm{s}} \prod_{u=k+1}^{\infty}\left(1-a s_u\right), \\
\end{aligned}
$$
where $h$ is a given large number that satisfies $s_k< 1/a,$ $\forall k \geq h$.\\
\\
Since Term (\ref{eq4.1}) $=$  Term (\ref{eq4}) $=0$ a.s.{},
$$
\begin{aligned}
&\sum_{k=h}^{\infty} \frac{\mu_n}{k^2 \sigma_n^2+1} \cdot \prod_{u=k+1}^{\infty} \frac{u^2 \sigma_n^2 (1-a s_u )}{u^2 \sigma_n^2+1}=0 \quad a.s.\\
&\prod_{k=h}^{\infty} \frac{k^2 \sigma_n^2\left(1-a s_k\right)}{k^2 \sigma_n^2+1} x_h=0 \quad a.s.{}
\end{aligned}
$$
\\
Moreover, by Lemma \ref{lem:important}, 
$$
\prod_{k=h}^{\infty}\left(1-a s_k\right) x_h^{\mathrm{s}}=0 \quad a.s.
$$
\\
Then, $\lim_{i \to \infty} x_{i+1}$ and $\lim_{i \to \infty} x_{i+1}^{\mathrm{s}}$ follow the same distribution if and only if 
\begin{align}
 \varepsilon_k^{\mathrm{s}}= \varepsilon_k \cdot \prod_{u=k}^{\infty} \frac{u^2 \sigma_n^2}{u^2 \sigma_n^2+1} \quad \forall k \in \mathbb{N^{+}}. \label{s1}
\end{align}
\\
In Eq.~(\ref{s1}), since $\{\varepsilon_k\}$ is bounded and i.i.d, $\{\varepsilon_k^{\mathrm{s}}\}$ is bounded and independent. Moreover, $E(\varepsilon_k)=E(\varepsilon_k^{\mathrm{s}})=0$ $\forall k \in \mathbb{N^{+}}$. Therefore, $\{x_i^{\mathrm{s}}\}$ is a standard RM sequence given that $\sum_{i=1}^\infty s_i=\infty$, $\sum_{i=1}^\infty s_i^2 < \infty$. Hence, $\{x_i^{\mathrm{s}}\}$ converges to $x_\textrm{t}$, and $\{x_i\}$ converges to $x_\textrm{t}$, which means that the prior-information Robbins--Monro sequence is convergent a.s.\ for linear $f(x)$ and normal prior distribution.
\end{proof}
\end{theorem}

\section{Convergence for nonlinear function $f$ and Gaussian prior} \label{s5}
Keep the prior distribution to be normal. When the underlying function $f$, e.g., in a root search, is a nonlinear continuously differentiable function, 
$$
x_{i+1} =\operatorname{argmax}_x\!\left(\mathcal{N}(x \mid \mu_n, \sigma_n^2) \cdot \mathcal{N}\Big(x \Bigm| x_i-s_i\big(f(x_i)-f(x_\textrm{t})+\varepsilon_i\big), \frac{1}{i^2}\Big)\right).
$$
Through the mean value theorem,
$$
f(x_i)-f(x_\textrm{t})=f^{\prime}(z_i)(x_i-x_\textrm{t}),
$$
where $z_i$ lies between $x_i$ and $x_\textrm{t}$.
Therefore,
$$
x_{i+1} =\operatorname{argmax}_x\!\left(\mathcal{N}(x \mid \mu_n, \sigma_n^2) \cdot \mathcal{N}\Big(x \Bigm| x_i-s_i(f^{\prime}(z_i)(x_i-x_\textrm{t})+\varepsilon_i), \frac{1}{i^2}\Big)\right).
$$
\\
\begin{theorem} \label{thm2.2}When the prior distribution is normal and $f(x)$ is nonlinear, 
$$\lim_{i \to \infty}x_{i+1}= x_\textrm{t} \quad a.s.$$
Therefore, the prior-information Robbins--Monro sequence is convergent a.s.\ for nonlinear $f(x)$ and normal prior distribution.
\begin{proof}
By exactly the same procedure as for Theorem \ref{thm1},
\begin{align}
\lim_{i \to \infty} x_{i+1}-x_\textrm{t} & =\prod_{k=h}^{\infty} \frac{k^2 \sigma_n^2 (1- s_k f^{\prime}(z_k) )}{k^2 \sigma_n^2+1} (x_h-x_\textrm{t} ) \label{eq11}\\
&+\sum_{k=h}^{\infty} \frac{ \mu_n-x_\textrm{t} }{k^2 \sigma_n^2+1} \cdot \prod_{u=k+1}^{\infty} \frac{u^2 \sigma_n^2 (1- s_u f^{\prime}(z_u) )}{u^2 \sigma_n^2+1} \label{eq11.1}\\
& +\sum_{k=h}^{\infty} \frac{k^2 \sigma_n^2 s_k \varepsilon_k}{k^2 \sigma_n^2+1} \cdot \prod_{u=k+1}^{\infty} \frac{u^2 \sigma_n^2 (1- s_u f^{\prime}(z_u) )}{u^2 \sigma_n^2+1}, \label{eq11.2}
\end{align}
where $h$ is a given large number that satisfies $s_k< 1/M,$ $\forall k \geq h$. \\
\\
Note that $M\geq f^{\prime} \geq m>0$.\\
\begin{align*}
|(\ref{eq11})| \leq \left|\prod_{k=h}^{\infty} \frac{k^2 \sigma_n^2}{k^2 \sigma_n^2+1}\left(1-s_k m\right)\left(x_h-x_\textrm{t}\right) \right|
\end{align*}
Replacing $a$ by $m$ in Term (\ref{eq4}) entails 
$$\prod_{k=h}^{\infty} \frac{k^2 \sigma_n^2 (1- s_k m )}{k^2 \sigma_n^2+1} (x_h-x_\textrm{t} ) \label{eq2nl}  = 0 \quad a.s.$$
Therefore, Term (\ref{eq11}) $  = 0$. Following the same idea, we can get
$$
0 \leq |(\ref{eq11.1})| \leq \left|\sum_{k=h}^{\infty} \frac{ \mu_n-x_\textrm{t} }{k^2 \sigma_n^2+1} \cdot \prod_{u=k+1}^{\infty} \frac{u^2 \sigma_n^2 (1- s_u m )}{u^2 \sigma_n^2+1} \right| =0 \quad a.s.{},
$$
$$
0 \leq E[(\ref{eq11.2})^2] \leq \sum_{k=h}^{\infty} \left(\frac{k^2 \sigma_n^2 s_k}{k^2 \sigma_n^2+1} \right)^2 d^2 \cdot \prod_{u=k+1}^{\infty} \left(\frac{u^2 \sigma_n^2 (1-m s_u )}{u^2 \sigma_n^2+1} \right)^2 =0 \quad a.s.
$$\\
Therefore, Term (\ref{eq11.1}) = Term (\ref{eq11.2}) $=0$ a.s.
$$\lim_{i \to \infty}x_{i+1} = x_\textrm{t} \quad a.s.$$
Therefore, the prior-information Robbins--Monro sequence is convergent a.s.\ for nonlinear $f(x)$ and normal prior distribution.
\end{proof}
\end{theorem}

\section{Convergence for practically arbitrary prior} \label{s6}
The underlying function $f(x)$ in this section will be assumed to be a nonlinear function satisfying Assumption \ref{a3}.

\subsection{Weighted sum of Gaussian prior}
In this section, before going to the arbitrary prior, we will first assume the prior distribution to be the weighted sum of $L$ normal distributions with different means and the same variance $\sigma^2$. Then, 
\begin{align}
x_{i+1}  =\operatorname{argmax}_x\! \left( \Big(\sum_{r=1}^L w_r  \mathcal{N}(x \mid \mu_r, \sigma^2 )\Big) \cdot \mathcal{N}\Big(x \Bigm| x_i-s_i(y_i-y_\textrm{t}), \frac{1}{i^2}\Big) \right),\label{eq16}
\end{align}
where $0<w_r \leq 1$ $\forall r \in \{1,2,...,L\}$ is the weight of the $r^{th}$ normal distribution. If we want $\sum_{r=1}^L w_r\mathcal{N}(x | \mu_r, \sigma^2 )$ to be a probability distribution, we need to have $\sum_{r=1}^L w_r =1$. Note that the solution of Eq.\ (\ref{eq16}) always exists. Then, let the solution of
\begin{align}
\operatorname{argmax}_x \!\left(\mathcal{N}(x \mid \mu_i^{\mathrm{A}}, \sigma^2) \cdot \mathcal{N}\Big(x \Bigm| x_i-s_i(y_i-y_\textrm{t}), \frac{1}{i^2}\Big)\right) \label{eq16.5}
\end{align}
also be $x_{i+1}$, which follows if $\mu_i^{\mathrm{A}}$ satisfies
\begin{align}
x_{i+1}=\frac{i^2 \sigma^2(x_i-s_i(y_i-y_\textrm{t}))+\mu_i^{\mathrm{A}}}{i^2 \sigma^2+1},\label{eq13.05}\\
\mu_i^{\mathrm{A}}=(i^2\sigma^2+1) x_{i+1}- i^2\sigma^2 (x_i-s_i(y_i-y_\textrm{t})). \label{eq13.1}
\end{align}
Note that $\mathcal{N}(x \mid \mu_i^{\mathrm{A}}, \sigma^2)$ in Eq.\ (\ref{eq16.5}) will change in each iteration (change with $i$).\\
\\
\begin{theorem}\label{thm2.3}
The prior-information Robbins--Monro sequence is convergent a.s.\ when the prior distribution is the weighted sum of a finite number of normal distributions with the same variance $\sigma^2$. The weights $w_r$ satisfy $0<w_r \leq 1$, $\forall r \in \{1,2,...,L\}$ and $\sum_{r=1}^L w_r =1$.
\begin{proof}
$$
\begin{aligned}
x_{i+1}&=\operatorname{argmax}_x\!\left( \Big(\sum_{r=1}^L w_r \mathcal{N}(x \mid \mu_r, \sigma^2 )\Big) \cdot \mathcal{N}\Big(x \Bigm| x_i-s_i(y_i-y_\textrm{t}), \frac{1}{i^2}\Big)\right)\\
& =\operatorname{argmax}_x \!\left(\mathcal{N}(x \mid \mu_i^{\mathrm{A}}, \sigma^2) \cdot \mathcal{N}\Big(x \Bigm| x_i-s_i(y_i-y_\textrm{t}), \frac{1}{i^2}\Big)\right).  \\  
\end{aligned}
$$
Firstly, define 
\begin{align}
z_r &=\operatorname{argmax}_x\!\left( \mathcal{N}(x \mid \mu_r, \sigma^2 ) \cdot \mathcal{N}\Big(x \Bigm| x_i-s_i(y_i-y_\textrm{t}), \frac{1}{i^2}\Big)\right) \notag\\
& = \frac{i^2 \sigma^2(x_i-s_i(y_i-y_\textrm{t}))+\mu_r}{i^2 \sigma^2+1}.\label{eq13.2}
\end{align}
Obviously, $x_{i+1}$ lies between $\min\{z_r \mid r=1,2,...,L\}$ and $\max\{z_r \mid r=1,2,...,L\}$. By substituting Eq.\ (\ref{eq13.2}) into Eq.\ (\ref{eq13.1}), we can derive that $\mu_i^{\mathrm{A}}$ lies between $\min\{\mu_r \mid r=1,2,...,L\}$ and $\max\{\mu_r \mid r=1,2,...,L\}$, and hence $\{\mu_i^{\mathrm{A}}\}$ is a bounded sequence. Then, expand Eq.\ (\ref{eq13.05}) as in Theorems \ref{thm1} and \ref{thm2.2}, we can get 
\begin{align}
\lim_{i \to \infty} x_{i+1}-x_\textrm{t} & =\prod_{k=h}^{\infty} \frac{k^2 \sigma_n^2 (1- s_k f^{\prime}(z_k) )}{k^2 \sigma_n^2+1} (x_h-x_\textrm{t} ) \label{eq17}\\
&+\sum_{k=h}^{\infty} \frac{ \mu_k^{\mathrm{A}} -x_\textrm{t} }{k^2 \sigma_n^2+1} \cdot \prod_{u=k+1}^{\infty} \frac{u^2 \sigma_n^2 (1- s_u f^{\prime}(z_u) )}{u^2 \sigma_n^2+1} \label{eq18} \\
& +\sum_{k=h}^{\infty} \frac{k^2 \sigma_n^2 s_k \varepsilon_k}{k^2 \sigma_n^2+1} \cdot \prod_{u=k+1}^{\infty} \frac{u^2 \sigma_n^2 (1- s_u f^{\prime}(z_u) )}{u^2 \sigma_n^2+1}, \label{eq19}
\end{align}
where $h$ is a given large number that satisfy $s_k< 1/M,$ $\forall k \geq h$. \\
\\
Following the same proof as for Theorem \ref{thm2.2}, we can see that Term (\ref{eq17}) = Term (\ref{eq19}) $= 0$ a.s.\ since Term (\ref{eq11}) = Term (\ref{eq11.2}) $= 0$ a.s.\ Let $|\mu_B-x_\textrm{t}|= \max\left\{ |\mu_r-x_\textrm{t}| \bigm| r=1,2,...,L\right\}$, $B \in \{1,2,...,L\}$. Then,
$$
0 \leq |(\ref{eq18})| \leq \left|\sum_{k=h}^{\infty} \frac{ \mu_B-x_\textrm{t} }{k^2 \sigma_n^2+1} \cdot \prod_{u=k+1}^{\infty} \frac{u^2 \sigma_n^2 (1- s_u m )}{u^2 \sigma_n^2+1} \right| =0 \quad a.s.
$$
since Term (\ref{eq11.1}) = Term (\ref{eq4.1}) $= 0$ a.s.\\
\\
Therefore, Term (\ref{eq18}) $= 0$ ~~a.s.{}, $$\lim_{i \to \infty}x_{i+1} = x_\textrm{t} \quad a.s.$$
The prior-information Robbins--Monro sequence is convergent a.s.\ when the prior distribution is the weighted sum of a finite number of normal distributions with the same variance $\sigma^2$.
\end{proof}
\end{theorem}

\subsection{Discrete arbitrary prior}

We want to extend our results to more arbitrary priors. We will use kernel density estimation (KDE) to attain this goal \cite{chen2017tutorial}. Firstly, assume the original prior information of $x_\textrm{t}$ $\big(P_{x_\textrm{t}}(x)\big)$ is discrete. In nature, all the primary measurement results we have are discrete since we cannot perform infinitely many measurements.  

Assume we have $L$ measurements of $f(x)$ (at $x_1^{\mathrm{M}}, x_2^{\mathrm{M}},...,x_L^{\mathrm{M}}$) with the result $y_\textrm{t}$, and we will use the normal kernel function 
$$
K(x) = \frac{1}{\sqrt{2\pi}} e^{-\frac{x^2}{2}}
$$
in our KDE analysis. Note that $\sigma$ in Theorem \ref{thm2.3} is exactly the bandwidth of KDE. Then, with these discrete measurement results, we will approximate the original prior distribution $P_{x_\textrm{t}}(x)$ by
\begin{align}
P_{0}(x)=\frac{1}{L \sigma} \sum_{r=1}^L K \!\!\left(\frac{x_j^{\mathrm{M}}-x}{\sigma}\right), \label{eqp}
\end{align}
which is the averaged sum of normal distributions with the same variance. Here, $w_r \equiv 1/L$. Therefore, for any discrete prior information of $x_\textrm{t}$, we can use KDE to pre-process the prior information to get the approximated continuous prior distribution $P_{0}(x)$ (Eq.\ (\ref{eqp})) and derive an a.s.\ convergent prior-information Robbins--Monro sequence.

\subsection{Continuous practically arbitrary prior}

Now, we will consider the case that the original prior distribution is positive, bounded, and continuous. We will need the following assumptions and notation:

\begin{assumption} \label{a7} 
$s_i=1/i$.
\end{assumption}

\begin{assumption} \label{a8} 
The function underlying the prior-information Robbins--Monro sequence $f(x)$ satisfies 
$$
c_2 x - d_2 < f(x) \leq y_\textrm{t}-\delta \text{,\quad for } x<x_\textrm{t},
$$
$$
c_1 x + d_1 > f(x) \geq y_\textrm{t}+\delta \text{,\quad for } x>x_\textrm{t},
$$
for some positive constants $c_1$, $c_2$, $d_1$, $d_2$, $\delta $.
\end{assumption}

\begin{assumption} \label{a9}
The prior distribution $P_{x_\textrm{t}}(x)$ is a finite continuous distribution that satisfies $|P_{x_\textrm{t}}^{\prime}(x)/ P_{x_\textrm{t}} (x)| \leq J$ for some $J>0$. This condition guarantees that $ P_{x_\textrm{t}}^{\prime}(x)$ is bounded. Furthermore, the prior should be bounded per $ 0< P_{x_\textrm{t}} (x) < \infty$. 
\end{assumption}

\begin{notation}
Denote the RM distribution at the $i^{th}$ iteration 
$$\mathcal{N}\Big(x \Bigm| x_i-s_i\big(f(x_i)-f(x_\textrm{t})+\varepsilon_i\big), \frac{1}{i^2}\Big)$$ 
by $N_i(x)$. The derivative of $N_i(x)$ is $N_i^{\prime}(x)$.
\end{notation}

\begin{lemma}\label{lemcts}
Assume Assumptions \ref{a7}--\ref{a9} hold. Let
$$
\begin{aligned}
x_{i+1}&=\operatorname{argmax}_x\big(P_{x_\textrm{t}}(x) \cdot N_i(x)\big),\\
\mu_i^{\mathrm{c}}&=x_i-s_i\big(f(x_i)-f(x_\textrm{t})+\varepsilon_i\big).
\end{aligned}
$$
Then,
$$|x_{i+1}-\mu_i^{\mathrm{c}}| \leq \frac{J}{i^2}.$$

\begin{proof}
$$
\begin{aligned}
 N_i(x)=&\mathcal{N}\bigg(x \biggm| x_i-s_i\big(f(x_i)-f(x_\textrm{t})+\varepsilon_i\big), \frac{1}{i^2}\bigg)\\
=&\frac{i}{\sqrt{2 \pi}} \exp\!\left(-\frac{i^2\Big(x-x_i+s_i\big(f(x_i)-f(x_\textrm{t})+\varepsilon_i\big)\Big)^2}{2}\right) \\
 N_i^{\prime}(x)=&N_i(x) \cdot(-i^2(x-x_i+s_i(f(x_i)-f(x_\textrm{t})+\varepsilon_i)))
\end{aligned}
$$

Since $\mu_i^{\mathrm{c}}=x_i-s_i(f(x_i)-f(x_\textrm{t})+\varepsilon_i)$, 
$$\frac{N_i^{\prime}(x)}{N_i(x)}=-i^2(x-\mu_i^{\mathrm{c}}).$$

When $N_i(x) \cdot P_{x_\textrm{t}}(x)$ attains its maximum at $x_{i+1}$, 
$$
x_{i+1}=\operatorname{argmax}_x\big(P_{x_\textrm{t}}(x) \cdot N_i(x)\big),
$$
$$
N_i^{\prime}(x_{i+1}) P_{x_\textrm{t}}(x_{i+1})+ N_i(x_{i+1}) P_{x_\textrm{t}}^{\prime}(x_{i+1})=0
$$ 
$$
\frac{N_i^{\prime}(x_{i+1})}{N_i(x_{i+1})}=-\frac{P_{x_\textrm{t}}^{\prime} (x_{i+1})}{P_{x_\textrm{t}}(x_{i+1})}.
$$

Note that 
$$\left|\frac{P_{x_\textrm{t}}^{\prime}(x_{i+1})}{P_{x_\textrm{t}}(x_{i+1})}\right| \leq J.$$

Therefore, 
$$
\left|\frac{N_i^{\prime}(x_{i+1})}{N_i(x_{i+1})}\right|=\left|i^2(x_{i+1}-\mu_i^{\mathrm{c}})\right| \leq J,
$$
$$|x_{i+1}-\mu_i^{\mathrm{c}}| \leq \frac{J}{i^2}.$$

\end{proof}
\end{lemma}

\begin{theorem}\label{thmcts}
When Assumptions \ref{a7}--\ref{a9} hold, the prior-information Robbins--Monro sequence is a.s.\ convergent. 

\begin{proof}
We will continue to use the notations in Lemma \ref{lemcts}. We have seen that introducing the prior distribution at the $i^{\text {th}}$ iteration will at most pull $x_{i+1}$ away from $\mu_i^{\mathrm{c}}$ by $J/i^2$. Note that $\mu_i^{\mathrm{c}}$ is the next step if the sequence follows the standard RM scheme (without the prior information).

Let $f^{\mathrm{s}}(x_i)$ be the underlying function of a potential standard RM sequence $\{x_{i+1}^{\mathrm{s}}\}$ such that $x_{i+1}^{\mathrm{s}}=x_{i+1}$ $\forall i \in \mathbb{N}$. We will prove that $\{x_{i+1}^{\mathrm{s}}\}$ is a standard RM sequence. Since
$$
\begin{aligned}
x_{i+1}^{\mathrm{s}}=x_{i+1}=&x_i-s_i(f^{\mathrm{s}}(x_i)-f(x_\textrm{t})+\varepsilon_i)\\
\in & \left[\mu_i^{\mathrm{c}}-\frac{J}{i^2},~ \mu_i^{\mathrm{c}}+\frac{J}{i^2}\right]\\
= & \left[x_i-s_i(f(x_i)-f(x_\textrm{t})+\varepsilon_i)-\frac{J}{i^2},~ x_i-s_i(f(x_i)-f(x_\textrm{t})+\varepsilon_i)+\frac{J}{i^2}\right]
\end{aligned}
$$
and $s_i =1/i$,
$$
f^{\mathrm{s}}(x_i) \in \left[f(x_i)-\frac{J}{i},~ f(x_i)+\frac{J}{i}\right].
$$

Hence, from the above equations, we may easily see that to let $x_{i+1}^{\mathrm{s}}=x_{i+1}$, that is, to let
$$
\operatorname{argmax}_x\big(P_{x_\textrm{t}}(x) \cdot N_i(x)\big)= x_i-s_i(f^{\mathrm{s}}(x_i)-f(x_\textrm{t})+\varepsilon_i),  
$$
$f^{\mathrm{s}}(x_i)$ is at most perturbed by $J/i$ relative to $f(x_i)$. 

Then, we will define $f^{\mathrm{s}}(x)$ as follows: 

If $x \neq x_i(= x_i^{\mathrm{s}}),$ $ \forall i \in \mathbb{N}^{+}$, or $x = x_i(=x_i^{\mathrm{s}})$, for some $i \leq 2J/\delta$, let $f^{\mathrm{s}}(x) = f(x)$. Note that we defined $\delta$ in Assumption \ref{a8}.

If $x = x_i(=x_i^{\mathrm{s}})$, for some $i > 2J/\delta$, the value of $f^{\mathrm{s}}(x_i)$ is chosen such that 
$$
\begin{aligned}
x_{i+1}^{\mathrm{s}}=x_i-s_i(f^{\mathrm{s}}(x_i)-f(x_\textrm{t})+\varepsilon_i)=x_{i+1} = \operatorname{argmax}_x\big(P_{x_\textrm{t}}(x) \cdot N_i(x)\big).
\end{aligned}
$$

Since the values of $x_1, x_2,...$ are influenced by a number of continuous random variables $\varepsilon_1, \varepsilon_2,...$, we can see that $\forall i,j \in \mathbb{N}^{+}$, $x_i \neq x_j$ a.s. Therefore, $f^{\mathrm{s}}(x)$ is well-defined here.

Since $f(x_{i+1}^{\mathrm{s}}) \in [f(x_i)-J/i,f(x_i)+J/i]$ for $i > 2J/\delta$, $f(x_{i+1}^{\mathrm{s}}) \in [f(x_i)-\delta/2,f(x_i)+\delta/2]$. Moreover, $f(x^{\mathrm{s}})=f(x)$ for the other points $x$. Therefore, 
$$
f^{\mathrm{s}}(x) \leq y_\textrm{t}-\delta/2 \text{,\quad for } x<x_\textrm{t},
$$
$$
f^{\mathrm{s}}(x) \geq y_\textrm{t}-\delta/2 \text{,\quad for } x>x_\textrm{t},
$$
given that 
$$
f(x) \leq y_\textrm{t}-\delta \text{,\quad for } x<x_\textrm{t},
$$
$$
f(x) \geq y_\textrm{t}-\delta \text{,\quad for } x>x_\textrm{t}
$$
by Assumption \ref{a8}.

Moreover, since $f(x)$ satisfies
$$
f(x) > c_2 x - d_2  \text{,\quad for } x<x_\textrm{t},
$$
$$
f(x) < c_1 x + d_1  \text{,\quad for } x>x_\textrm{t},
$$
for some positive constants $c_1$, $c_2$, $d_1$, $d_2$ by Assumption \ref{a8}, we can easily see that $f^{\mathrm{s}}(x)$ satisfies
$$
f^{\mathrm{s}}(x) > c_3 x - d_3  \text{,\quad for } x<x_\textrm{t},
$$
$$
f^{\mathrm{s}}(x) < c_4 x + d_4  \text{,\quad for } x>x_\textrm{t},
$$
for some positive constants $c_3$, $c_4$, $d_3$, $d_4$ since the difference between $f^{\mathrm{s}}(x)$ and $f(x)$ is very small.

By Blum, the underlying function $f(x)$ of the standard RM sequence needs to satisfy the following conditions \cite{blum1954approximation}:
$$
c_2 x - d_2 < f(x) < y_\textrm{t} \text{,\quad for } x<x_\textrm{t},
$$
$$
c_1 x + d_1 > f(x) > y_\textrm{t} \text{,\quad for } x>x_\textrm{t}.
$$

Note that $\{\varepsilon_i\}$ have finite variance. 
Therefore, by Blum's argument, $f^{\mathrm{s}}(x)$ can be the underlying function of a standard RM sequence. $\{x_{i}\}$ = $\{x_i^{\mathrm{s}}\}$ are a.s.\ convergent standard RM sequences when $i > 2J/\delta$. These two sequences are generated by $x_{i+1}= x_{i+1}^{\mathrm{s}}=x_i^{\mathrm{s}}-s_i(f^{\mathrm{s}}(x_i)-f(x_\textrm{t})+\varepsilon_i)$.

When $i \leq 2J/\delta$, we just simply make $x_i^{\mathrm{s}}=x_i= \operatorname{argmax}_x\!\big(P_{x_\textrm{t}}(x) \cdot N_i(x)\big)$. Note that the initial few steps will not influence the convergence of the whole sequence. It will not influence whether $\{x_i^{\mathrm{s}}\}$ is a standard RM sequence or not.

Therefore, with Assumptions \ref{a7}--\ref{a9}, the prior-information Robbins--Monro sequence $\{x_{i}\}$ is indeed a standard RM sequence. Therefore, it is a.s\ convergent. 

\end{proof}
\end{theorem}

\begin{remark}
Note that we require the original prior distribution $P_{x_\textrm{t}}(x)$ to be always positive and bounded, i.e., $0< P_{x_\textrm{t}}(x)<\infty$. Otherwise, the prior-information Robbins--Monro sequence may not converge to $x_\textrm{t}$. 
\end{remark}

\begin{remark}
We may see that if $P_{x_\textrm{t}}(x)$ is a polynomial in a closed interval $[a_0,a_1]$ and satisfies $ \inf_{x \in [a_0,a_1]} P_{x_\textrm{t}}(x)>0$, the Assumption \ref{a9} will be satisfied for $P_{x_\textrm{t}}(x)$ locally in that interval.

\end{remark}

Afterwards, we will go back to use $P_0(x)$ (in Eq.\ (\ref{eqp})) to estimate the continuous positive bounded prior distribution $P_{x_\textrm{t}}(x)$. The KDE $P_0(x)$ has several important properties. These properties are largely a result of the characteristics of the normal kernel and the non-parametric nature of KDE \cite{higgins2004introduction}:

(1) Smoothness: $P_0(x)$ produces a smooth density estimate. The Gaussian kernel is infinitely differentiable, which means that the resulting density estimate is also smooth and has no discontinuities. 

(2) Non-Negative: The estimated density function is always non-negative. 

(3) Normalisation: The integral over the entire space of $P_0(x)$ equals 1, satisfying the property of a probability density function. 

(4) Bandwidth Dependence: The choice of bandwidth ($\sigma$) has a significant effect on $P_0(x)$. A small bandwidth leads to a density estimate that is more sensitive to fluctuations in the data (high variance but low bias), potentially resulting in an overfit. A large bandwidth smooths out these fluctuations (low variance but high bias), which might oversimplify the structure of the data.

(5) Pointwise Convergence: As the sample size increases and thus the number of Gaussians in the series, $P_0(x)$ converges to $P_{x_\textrm{t}}(x)$ at each point.

(6) Asymptotic Properties: $P_0(x)$ has well-defined asymptotic properties. As the number of data points increases, $P_0(x)$ approaches $P_{x_\textrm{t}}(x)$, provided the bandwidth is chosen appropriately.

(7) Boundedness: From Eq.\ (\ref{eqp}), we can see that $0<P_0(x) \leq 1/(\sqrt{2 \pi}\sigma)$, where the bandwidth $\sigma$ determines its maximum. Moreover, given the bandwidth $\sigma$, $P_0(x)$ is always finite. Note that the spread of $x_1^{\mathrm{M}}, x_2^{\mathrm{M}},...,x_L^{\mathrm{M}}$ will also influence the maximum of $P_0(x)$. If they are centralised, the maximum of $P_0(x)$ will be larger. If they are decentralised, the maximum of $P_0(x)$ will be lower.

(8) Handling Multimodal Distributions: $P_0(x)$ is capable of representing multimodal distributions (distributions with multiple peaks). 

(9) Local Property: The estimation at each point is influenced mainly by nearby data points due to the local nature of the Gaussian kernel. This property allows $P_0(x)$ to adapt to the local structure of the data.

(10) Global Property: The Gaussian kernel, with its infinite support, means that every data point has some influence on every estimation point, which is beneficial for capturing global structures.

(11) Robustness and Flexibility: $P_0(x)$ does not make strong assumptions about $P_{x_\textrm{t}}(x)$, rendering it particularly useful in situations where $P_{x_\textrm{t}}(x)$ is unknown or complicated.

(12) Kernel Symmetry: The Gaussian kernel is symmetric, which simplifies the mathematical treatment of $P_0(x)$ and ensures that the density estimate is not skewed in any particular direction around a data point.

If we draw $L$ random discrete samples from $P_{x_\textrm{t}}(x)$ to estimate $P_{x_\textrm{t}}(x)$ by $P_0(x)$ (in Eq.\ (\ref{eqp})), the approximated prior distribution $P_{0}(x)$ is always positive and finite. Hence, the prior-information Robbins--Monro sequence with $P_{0}(x)$ (as the prior distribution) is always convergent. If the original continuous prior distribution $P_{x_\textrm{t}}(x)=0$ or $\infty$ somewhere, without pre-processing it, the corresponding prior-information Robbins--Monro sequence may not converge to $x_\textrm{t}$.

By the Law of Large Numbers, when the sample size $L$ is sufficiently large, the distribution of the discrete samples will be very close to the original continuous prior distribution $P_{x_\textrm{t}}(x)$. Higgins showed that the order of mean integrated square error (MISE) between $P_{0}(x)$ and $P_{x_\textrm{t}}(x)$ is $\mathcal{O}(\sigma^4)+\mathcal{O}(1/L\sigma)$ \cite{higgins2004introduction}. Therefore, we can see that when $\sigma$ is close to 0 ($\sigma>0$) and $L \gg 1/\sigma$, the MISE between $P_{0}(x)$ and $P_{x_\textrm{t}}(x)$ will be very small. Hence, the KDE $P_{0}(x)$ with a small bandwidth ($\sigma$) and large sample size ($L$) will be a good estimate of $P_{x_\textrm{t}}(x)$. Since the prior-information Robbins--Monro sequence with $P_{0}(x)$ as the prior distribution is a.s.\ convergent according to Theorem \ref{thm2.3}, this sequence can be an estimate of the prior-information Robbins--Monro sequence with any prior distribution $P_{x_\textrm{t}}(x)$ of practical concern if $0< P_{x_\textrm{t}}(x)<\infty$.

Moreover, Alspach and Sorenson previously supported that the Gaussian sum in the form of 
\begin{align}
P_A(x) = \sum_{r=1}^L w_r N(x \mid a_r, \sigma^2) \label{als}
\end{align}
(where $\sum_{r=1}^L w_r=1$, $w_r \geq 0$ for all $r$) converges uniformly to any probability density function of practical value as the number of terms $L$ increase and the variance $\sigma^2$ approaches 0 \cite{alspach1972nonlinear}. If $\sigma$ is small enough but nonzero, any positive finite continuous prior distribution $P_{x_\textrm{t}}(x)$ can be approximated by a Gaussian sum in the form of Eq.\ (\ref{als}) to any accuracy. Moreover, Sorenson and Alspach further showed that if $P_{x_\textrm{t}}(x)$ is continuous at all but a finite number of locations, we can still use $P_\textrm{A}(x)$ to approximate $P_{x_\textrm{t}}(x)$ \cite{sorenson1971recursive}. Moreover, we proved that the prior-information Robbins--Monro sequence, whose prior distribution is a convex combination of same-variance Gaussians, is a.s.\ convergent per Theorem \ref{thm2.3}. In combination, the prior-information Robbins--Monro sequence is a.s.\ convergent for all practical positive finite prior distributions $P_{x_\textrm{A}}(x)$, which well approximate any prior distribution $P_{x_\textrm{t}}(x)$ that is continuous at all but a finite number of points.

\subsection{More general convergence} \label{s7}

The above proof covers most practical prior distributions as it includes all priors that can be represented by a Gaussian kernel density estimate. We could alternatively assume any bounded and (beyond a finite number of discontinuous steps) continuous prior $P(x)$ greater than 0 so that the target $x_\textrm{t}$ and also any point between every $x_i$ and the target $x_\textrm{t}$ can be reached by the sequence. The boundedness excludes concepts such as Dirac distributions, which would derail the current sequence as it was defined and would need a modified formalism.
Convergence is guaranteed if a standard RM sequence $\{x^s_i\}$ exists that reproduces the exact same sequence of $\{x_i\}$ with a control sequence $\{s^s_i\}$ according to
\begin{align}
x^s_{i+1} = x^s_i - s^s_i(y^s_i - y_\textrm{t}), \\
x^s_i = x_i \quad \forall i, \nonumber \\
y^s_i = y_i \quad \forall i, \nonumber
\end{align}
so that
\begin{align}
\sum s^s_i = \infty, \label{cond_sdash1}\\
\sum (s^s_i)^{2} < \infty. \label{cond_sdash2}
\end{align}
Please note that with a prior distribution unequal to a constant, $\{s^s_i\}$ does not coincide with $s_i$ from the conventional RM sequence. Instead, $\{s^s_i\}$ incorporates the effect of the prior initially pulling the sequence closer to its own centre of gravity.
The existence of ${s^s_i}$ is trivial as it can be systematically constructed. For proving convergence, this sequence has to fulfill Conditions (\ref{cond_sdash1}) and (\ref{cond_sdash2}) in the long run.

\section{Numerical analysis} \label{s8}
\begin{notation}
$X=Lognormal(\alpha, \beta^2)$ is a random variable such that $\textrm{ln} X$ follows the normal distribution $\mathcal{N}(x \,|\, \alpha, \beta^2)$ with $\alpha$ mean and $\beta^2$ variance.
\end{notation}
\begin{notation}
$X=10^{\mathcal{U}(\alpha, \beta)}$ is a distribution such that $\textrm{lg}X = \textrm{log}_{10} X$ follows the uniform distribution $\mathcal{U} (\alpha, \beta)$.
\end{notation}
We performed all numerical analysis in MATLAB R2016b (9.1.0.441655) 64-bit (win64). 
A key objective of this section is to find the optimal $c_{i=0}$ in 
\begin{align*}
x_{i+1} =\operatorname{argmax}_x\!\Big(\mathcal{N}(x \mid \mu_n, \sigma_n^2) \cdot \mathcal{N}\big(x \bigm| x_i-s_i(a(x_i-x_\textrm{t})+\varepsilon_i), c_i^2\big)\Big) ,
\end{align*}
\\
where $c_i=c_0/i$, $c_0$ is a positive constant, so that  the prior-information Robbins--Monro sequence converge fastest. The expectation of $\varepsilon_i$ is 0 and the standard deviation of $\varepsilon_i$ is $d$, that is, $\varepsilon_i = \mathcal{N}(0, d^2)$.

We will first find the parameters related to the optimal value of $c_0$ and demonstrate that the ${c_i}$ sequence should reflect the uncertainty of the observation of the function $f$. Then, we will give an equation describing the relationship between the optimal $c_0$ with these parameters by linear regression. Afterwards, we will find out how much the prior-information Robbins--Monro sequence with the optimal $c_0$ performs better than the standard RM algorithm.

In this section, the prior distribution is constantly $\mathcal{N}(0.5, (0.25)^2)$, and the underlining function of the RM algorithm is $f(x)=ax$. For each run (not each step/iteration) of the (standard or prior) RM algorithm, $a$ will be randomly chosen again according to $Lognormal(0,(0.5)^2)$ and $x_\textrm{t}$ will be randomly chosen under $\mathcal{N}(0.5, (0.25)^2)$. The step size gain sequence $s_i=s_1/i$, where $s_1$ is randomly chosen from $10^{\mathcal{U}(0, 1)}$. For each time, we want to measure $f(x_i)$, the result we get is $y(x_i) = y_i = f(x_i) + \varepsilon_i$. Since we will only run the algorithms for a finite number of iterations and repetitions, $\varepsilon_i$ can be seen as a bounded random variable in our simulation. For all the simulations below, we always run the whole algorithm for many times. All the results we show in this section will be the median (or the average of medians) of a number of runs (for example, the median deviation (of 400,000 runs) of the prior-information Robbins--Monro sequence at the $10^{th}$ iteration).

\begin{figure}
    \centering
    \includegraphics[scale=0.88]{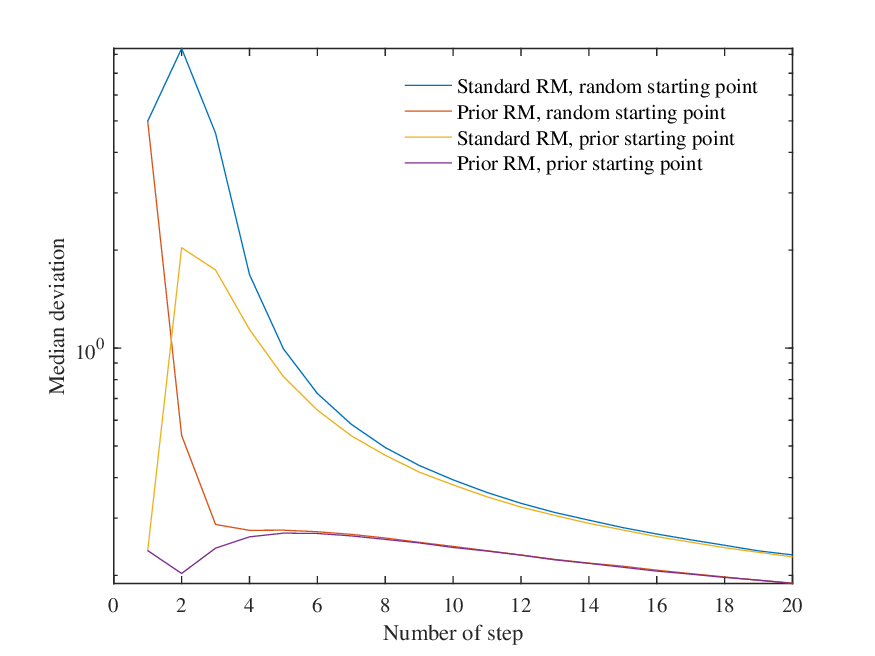}
    \caption{The median deviation ($ |x_i-x_\textrm{t}| $ or $|x_i^{\mathrm{s}}-x_\textrm{t}|$) of 400,000 runs for four algorithms in the first 20 steps ($c_0$ =0.3, $d$ =1).}
    \label{f1}
\end{figure}

Four algorithms are investigated in Figure \ref{f1}. The algorithms with prior starting points will choose their starting points according to the prior distribution $ \mathcal{N}(0.5, (0.25)^2)$. The algorithms with naive random starting points will choose their starting points according to a uniform distribution of $\mathcal{U}(-10, 10)$. For each algorithm, we will investigate the deviation ($ |x_i-x_\textrm{t}| $ or $|x_i^{\mathrm{s}}-x_\textrm{t}|$) when $i=1, 2,..., 20$.

Figure \ref{f1} illustrates that the prior-information Robbins--Monro sequence converges faster than the standard RM, especially in the first few iterations and when $x_1$ (or $x_1^{\mathrm{s}}$) is far from $x_\textrm{t}$. Subsequently, we are naturally concerned about the parameters related to the optimal value of $c_0$ that let the prior-information Robbins--Monro sequence converge fastest.

\begin{figure}
    \centering
    \includegraphics[scale=0.62]{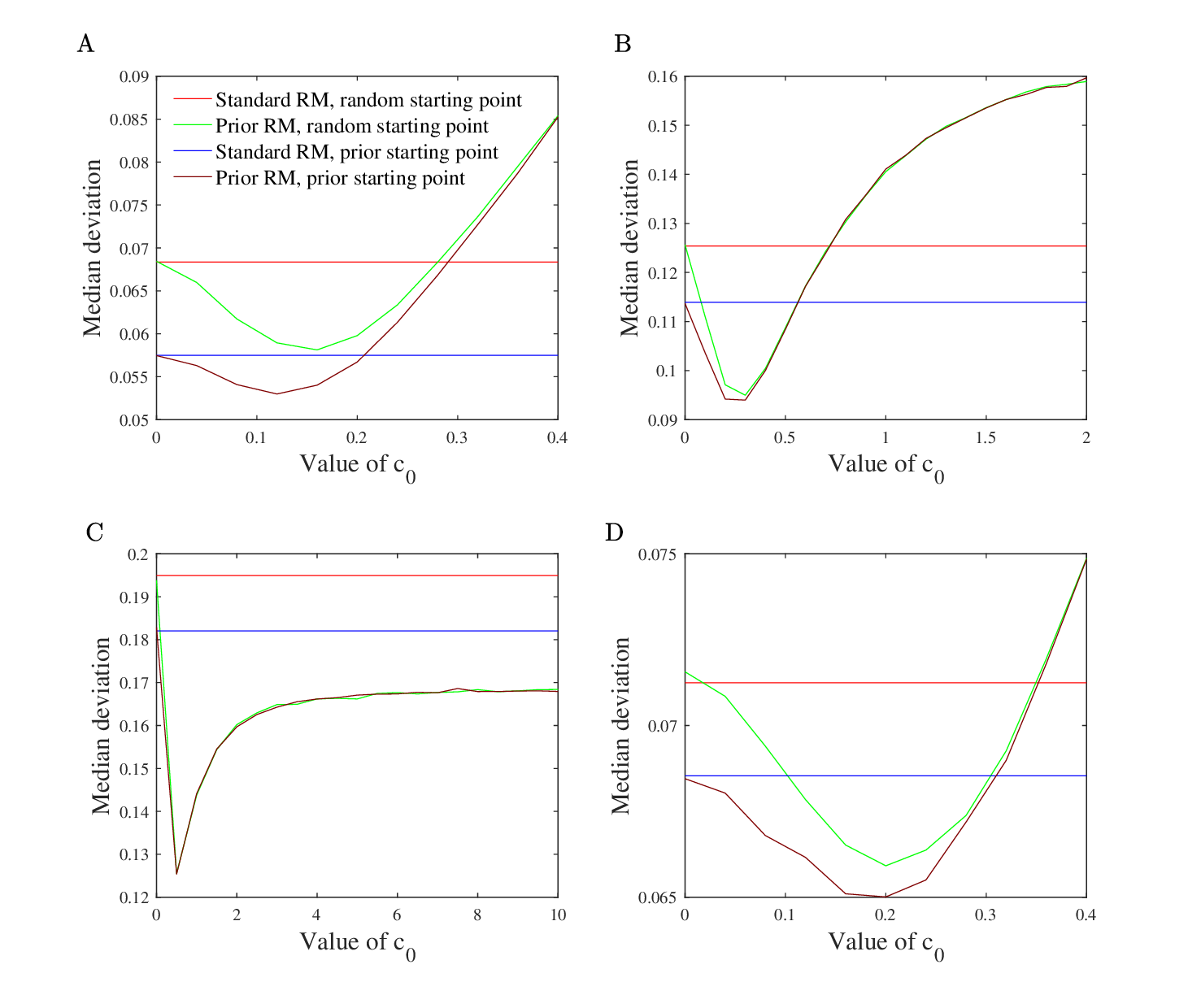}
    \caption{The median deviation ($ |x_i-x_\textrm{t}| $ or $|x_i^{\mathrm{s}}-x_\textrm{t}|$) of 400,000 runs at the $10^{th}$ step for the four algorithms when A $d=0.15$, B $d=0.3$, C $d=0.48$, and D at $20^{th}$ iteration when $d=0.3$.}
    \label{f2}
\end{figure}

In Figure \ref{f2} A, the reader may readily recognise that the prior-information Robbins--Monro sequence with $d=0.15$ performs best around the $10^{th}$ iteration when $c_0$ lies around $0.12$. From $c_0=0.12$ on and with larger $c_0$, the residual deviation of the prior-information Robbins--Monro sequence at the $10^{th}$ iteration increases, even exceeding the deviation of the standard RM when $c_0 > 0.3$.

An intuitive explanation for this phenomenon is a trade-off between the prior distribution and the noise contribution or uncertainty of the function observation. If $d$ is small, i.e., when the measurement of $f(x)$ is more accurate and less variable, a large value $c_0$ will spread out the RM distribution and flatten its maximum so that the prior distribution's influence dominates and pulls $x_{i+1}$ too close to $\mu_n$. Thus, the prior distribution would dominate. However, typically $\mu_n \neq x_\textrm{t}$. Then $x_{i+1}^{\mathrm{s}}$ may be closer to $x_\textrm{t}$ than $x_{i+1}$ when $c_0$ is too large.

Comparing Figure \ref{f2} A and B, the error of measuring $f(x)$ is doubled. Moreover, the optimal value of $c_0$ also roughly doubles, increasing from around $0.12$ to $0.3$. When the variability $d$ of the observation increases, $x_{i+1}$ and $x_{i+1}^{\mathrm{s}}$ naturally converge slower as this variability has to be averaged out over more iterations. An adjustment of the variance of the RM distribution automatically increases the influence of the prior distribution (increase $c_0$) and places $x_{i+1}$ closer to $\mu_n$ already in the first few iterations. The prior-information RM consequently accelerates the convergence since $\mu_n$ is an estimate of $x_\textrm{t}$. Again, however, the influence of the prior distribution may not be too large in this balance. Otherwise, it will take time until the accumulated information in the RM distribution (and the noise averaged out with every iteration) requires too many iterations before it outbalances the prior. 

If $d \geq 0.48$ (see Figure \ref{f2} C), the residual deviation of the prior-information Robbins--Monro sequence is always smaller than that of the standard RM at the $10^{th}$ iteration for all studied $c_0$. Hence, for large observation variability $d$, the prior-information Robbins--Monro sequence consistently outperforms standard RM.

Figure \ref{f2} D displays the residual deviation at the $20^{th}$ iteration when $d=0.3$. The optimal $c_0$ value appears to be around 0.2. Comparing Figures \ref{f2} B and D, we can see that the best-performing $c_0$ for the prior-information Robbins--Monro sequence also depends on the number of steps into the iteration. When the number of steps increases, the optimal $c_0$ tends to decrease.

\begin{figure}
    \centering
    \includegraphics[scale=0.8]{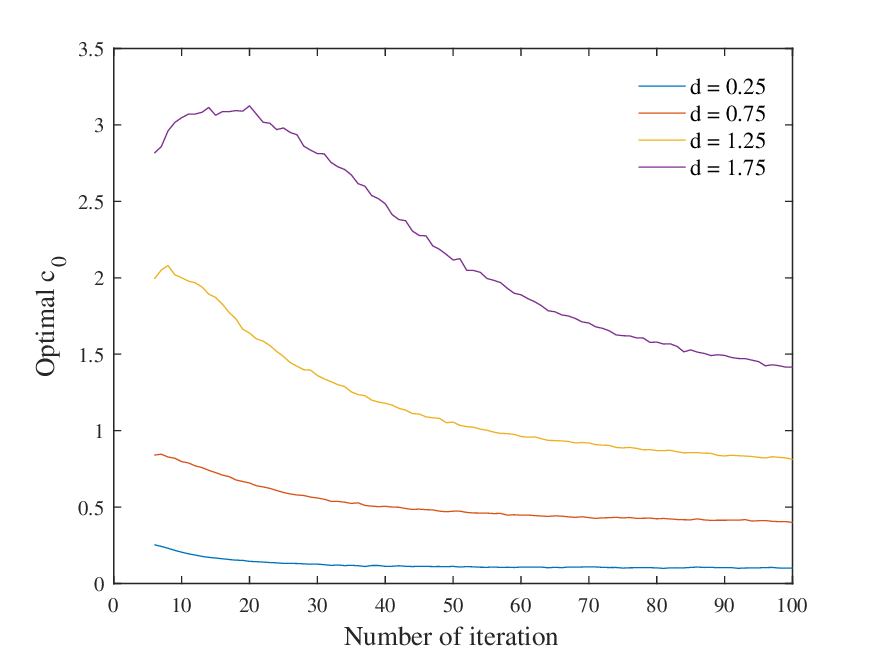}
    \caption{The graph of the optimal $c_0$ of the prior distribution for $d \in \{0.25, 0.75, 1.25, 1.75\}$ and iterations 6--100. Optimal $c_0$ for each $d$ and iteration is the average of 700 medians derived from 6,000 runs. (run the whole algorithm for $700 \times 6,000$ times)}
    \label{f3}
\end{figure}

In conclusion, the optimal initial variance $c_0$ of the RM distribution is related to the observation variability $d$ and the number of steps into the iteration. We identified the optimal $c_0$, i.e., the one with the smallest residual deviation from the convergence point, for each $d \in [0,2]$ over 6--100 iterations. In this numerical simulation, we chose the starting point of the iteration according to the prior distribution $\mathcal{N}(0.5, (0.25)^2)$. The first five iterations are excluded as these steps are highly unstable and rather random, so that they do not yet contain much information. A linear regression leads to
\begin{align}
c_0=1.32 \times d-0.0089 \times \mathrm{<Iteration>} +0.13. \label{c0}
\end{align}

The root-mean-squared error of the linear regression is 0.317, and $R^2=0.865$. The F-statistic vs.\ a constant model is 61400, and $p \ll 10^{-10}$. The linear regression result reflects the above-identified increase of the optimal $c_0$ with increasing observation variability $d$ and a minor decrease with the number of iterations. We presented the $c_0$--iteration relationship curve for $d \in \{ 0.25, 0.75, 1.25, 1.75\}$ in Figure \ref{f3}.

\begin{figure}
    \centering
    \includegraphics[scale=0.8]{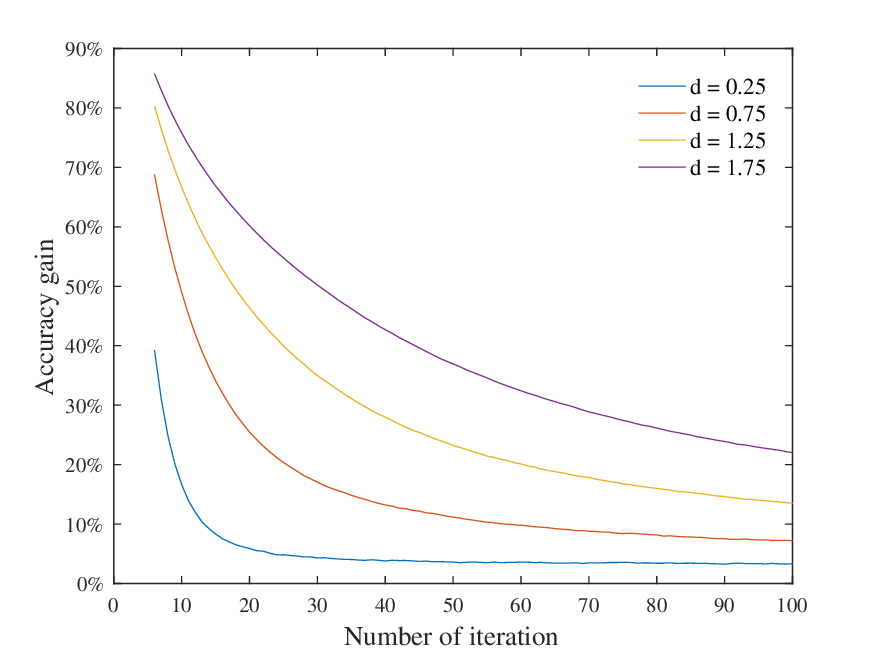}
    \caption{The graph of the performance of the prior-information Robbins--Monro sequence with the optimal $c_0$ for $d \in \{0.25, 0.75, 1.25, 1.75\}$ and iterations 6--100. Accuracy gain for each $d$ and iteration is the average of 700 medians derived from 6,000 runs. (run the whole algorithm for $700 \times 6,000$ times)}
    \label{f4}
\end{figure}

We further investigate the performance of the prior-information Robbins--Monro sequence with the optimal $c_0$. Then, we will run the RM algorithm (with and without prior) and stop at the iteration we designated to see the performance of the prior-information Robbins--Monro sequence at that iteration with the given optimal $c_0$. The results are shown in Figure \ref{f4}. Given the function observation variability $d$, we denote the standard and prior-information Robbins--Monro sequences' deviation at the $i^{th}$ iteration ($|x_i^{\mathrm{s}}-x_\textrm{t}|$ and $|x_i-x_\textrm{t}|$) respectively by $d_i^{\mathrm{s}}$ and $d_i$. The value in Figure \ref{f4} is the ratio of $(d_i^{\mathrm{s}}- d_i)/d_i^{\mathrm{s}}$ and thus reflects the gain of the prior-information RM over the standard RM. The starting points of both algorithms are chosen randomly according to the prior distribution $\mathcal{N}(0.5, (0.25)^2)$, i.e., the prior distribution.

Results in Figure \ref{f4} illustrate that for any observation variability $d$, the prior-information Robbins--Monro sequence will perform better than the standard RM sequence with the corresponding optimal $c_0$. After many iterations, the advantage washes out though. The prior-information Robbins--Monro sequence outperforms the standard RM primarily early into the iteration and/or for large function observation variability $d$.

\section{Conclusion} \label{s9}
This paper proposed a novel way to improve the convergence speed of the Robbins--Monro algorithm by turning it into a sequence of distributions and introducing prior information about the target point $x_\textrm{t}$. We proved the convergence of the prior-information Robbins--Monro sequence when the prior distribution is Gaussian or an averaged sum of same-variance Gaussians. The latter can approximate most practical arbitrary prior distributions using kernel density estimation with a normal kernel to derive an a.s.\ convergent prior-information Robbins--Monro sequence. We also gave the convergence proof of the prior-information Robbins--Monro sequence with an arbitrary continuous prior under some restrictions.

We further analysed the sequence for linear $f(x)$ with a random slope, which should represent the most practical functions locally. In the numerical results, the prior-information Robbins--Monro sequence consistently outperforms the standard Robbins--Monro iteration, especially in the first iterations when $x_1$ is far from $x_\textrm{t}$ and/or when the function's observation variability $d$ is large. The optimal initial spread of the RM distribution ($c_0$) is tightly linked to the random measurement error $d$ of $f(x)$. A smaller correction can refine this value depending on the intended number of iterations. A regression curve derived from our analysis can inform the best choice of $c_0$. 

A multidimensional prior-information Robbins--Monro sequence and the continuous-time prior-information Robbins--Monro sequence still require further research.

\section*{Author contributions}
SMG conceived, supervised, secured funding for the study, and provided the sequence for introducing prior information into the RM iteration. SMG also provided part of the code. SMG and SL conceived and sketched concepts for the proofs. SL completed all the proofs in this paper. SL further studied and characterised the sequence and its performance. SMG and KM checked the proofs. SL and KM also wrote the rest of the code for numerical analysis and visualisation. SL and SMG wrote the manuscript, KM edited the text and figures, and all authors reviewed, commented on, and approved the final version of the manuscript.

\bibliographystyle{imsart-number} 
\bibliography{bibliography}

\end{document}